\documentclass[conference]{IEEEtran}
\IEEEoverridecommandlockouts
\usepackage{cite}
\usepackage{amsmath,amssymb,amsfonts}
\usepackage{algorithmic}
\usepackage{graphicx}
\usepackage{textcomp}
\usepackage{comment}
\usepackage{booktabs}
\usepackage{mathtools}
\usepackage{hyperref}
\usepackage{subfigure}
\usepackage{caption}
\usepackage{ragged2e}
\usepackage{wrapfig}
\usepackage{lipsum}
\def\BibTeX{{\rm B\kern-.05em{\sc i\kern-.025em b}\kern-.08em
    T\kern-.1667em\lower.7ex\hbox{E}\kern-.125emX}}
    
\newcommand{\RN}[1]{%
  \textup{\uppercase\expandafter{\romannumeral#1}}%
}

\begin{document}

\title{QUARC: Quaternion Multi-Modal Fusion Architecture For Hate Speech Classification\\
{\footnotesize }
\thanks{\IEEEauthorrefmark{1} Authors have equal contribution}
}

\author{\IEEEauthorblockN{Deepak Kumar\IEEEauthorrefmark{1}\IEEEauthorrefmark{2},
Nalin Kumar\IEEEauthorrefmark{1}\IEEEauthorrefmark{3} and Subhankar Mishra\IEEEauthorrefmark{4}}
\IEEEauthorblockA{School of Computer Sciences, NISER, Bhubaneswar $752050$, India \\
Homi Bhabha National Institute, Training School Complex, Anushakti Nagar, Mumbai $400094$, India\\
Email: [\IEEEauthorrefmark{2}deepak.kumar, \IEEEauthorrefmark{3}nalin.kumar, \IEEEauthorrefmark{4}smishra]@niser.ac.in}}


\maketitle

\begin{abstract}
Hate speech, quite common in the age of social media, at times harmless but can also cause mental trauma to someone or even riots in communities. Image of a religious symbol with derogatory comment or video of a man abusing a particular community, all become hate speech with its every modality (such as text, image, and audio) contributing towards it. Models based on a particular modality of hate speech post on social media are not useful, rather, we need models like multi-modal fusion models that consider both image and text while classifying hate speech. Text-image fusion models are heavily parameterized, hence we propose a quaternion neural network-based model having additional fusion components for each pair of modalities. The Model is tested on the MMHS150K twitter dataset for hate speech classification. The model shows an almost $75\%$ reduction in parameters and also benefits us in terms of storage space and training time while being at par in terms of performance as compared to its real counterpart.
\end{abstract}


\section{Introduction}
Internet, social media, in particular, make a large and integral part of our society. Laws of civilized society apply there too but a certain sense of anonymity on social media brings the worst in many people. This results in incidents of hate speech in terms of online posts on social media. Hence, some moderation system is required to identify what can be classified as hate speech. To name a few websites doing such moderation are Facebook and Reddit. Due to volume, variety, and velocity of social media content \cite{ref_article14}, it renders human moderators useless, leading to the need for machine moderators (using machine learning). \par

Existing hate speech detection models in the text have achieved significant accuracy using standard architectures like SVM, MLP, DNN \cite{ref_article15}. However, indirect comments with some context can also be considered as hate speech. Also, with the rise of memes and emojis, hate speeches have become quite diverse and with it, the need for investigation in machine learning-based models for hate speech detection in images \cite{ref_article16} arises. Hence, one needs to consider all aspects of a social media post. However, such models could easily have parameters in millions \cite{ref_article5}, which poses its own difficulties in terms of practical deployment. This brings us to quaternion algebra, an extension of complex algebra. Recently its use in deep neural network has shown faster convergence and reduced parameters  \cite{ref_article4}. This rekindled interest in its possible application in machine learning. Quaternion neural networks have recently been considered in natural language processing (NLP) \cite{ref_article3}  tasks especially because of their ability to reduce parameters by $75\%$.\par

This paper introduces multi-modal fusion model of image, text, and text of image for hate speech classification on social media posts. We present a quaternion convolution neural network (QCNN) based fusion model and its simpler variants to do sentiment analysis for hate speech detection. Our model is inspired from existing symmetric gated fusion model \cite{ref_article5}. We apply QCNN on text, image, and image text separately to get three vectors. We then find additional fusion vectors which are quaternion attention on pairs of three modalities followed by their concatenation. The vector we finally get is used for hate speech classification. The model achieves expected $75\%$ parameter reduction and faster model training and also achieves better or similar precision scores than its real counterpart. \par

The main contributions of our paper are as follows:
\begin{itemize}
    \item Novel quaternion multi-modal fusion model for hate speech classification.
    \item Almost $75\%$ reduction in parameter size (hence, requires lesser storage space) of multi-modal fusion model as compared to current state of art.
    \item At par performance of quaternion multi-modal fusion model in comparison to its real counterpart.
\end{itemize}

\section{Related Works}
\label{sec:2}
Hate speech classification has been of interest for almost three decades. One of the earliest works in the hate speech domain was Ellen Spertus's Smokey \cite{ref_article6}, which classified private messages between two parties using a tree-based classifier. This paper used a rigid set of rules to create the feature vector. This model had several limitations, like, it could not handle unusual typography, grammatical mistakes or sarcasm. Though its performance was at par with that of human annotators' but what constitutes hate speech was also quite restrictive there.
\cite{ref_article7}  provides the hate speech classification as an NLP task on Web$2.0$ (The social media) using SVM classifier public forum chats. This model, in addition to the rule-based features as in Smokey, also has sentiment and context features. \cite{ref_article8} provides an extensive review of hate speech classification literature in terms of the NLP task and presents different data-sets available \& different approaches to the problem.
\par

With the ease of sharing different types of content over social media the modalities of hate speech also increased. One prominent example can be memes and emojis which contain image, text in the image, whose investigation requires a new approach as presented in \cite{ref_article16} which simply concatenated the different modalities. Another interesting work in multi-modal area is the tensor fusion network, a novel model for multi-modal sentiment analysis given by \cite{ref_article9} exploring inter and intra-modality associations. This model observed video data giving text, image, and audio modalities and fused them using vector product of all three. Further improvement was the inter-modal sentiment analysis using attention architecture over the tensor fusion approach by \cite{ref_article10}. In state-of-the-art \cite{ref_article5}, the authors proposed a series of multi-modal fusion models based on CNN, ResNet, and attention for hate speech classification on the Facebook data set. They claimed that an additional fusion vector improved accuracy. However, the number of parameters involved in the fusion model was too high.\par

Quaternion neural networks have been in computer science for quite some time in fields like signal processing and computer vision. Recently, using the quaternion transformer model \cite{ref_article3}, the authors showed that for almost all NLP tasks, quaternion model reduced $75\%$ parameters without any significant loss in accuracy. The reason was claimed to be the Hamilton product in quaternion algebra which is the counterpart of the dot product in real algebra. \cite{ref_article4} proposed QCNN, quaternion weight initialization, and batch normalization scheme. They experimented over the image dataset and showed that the quaternion model performed at par with the real model while reducing parameters and converging faster.

\begin{figure}[htbp]
\centerline{\includegraphics[width=\linewidth]{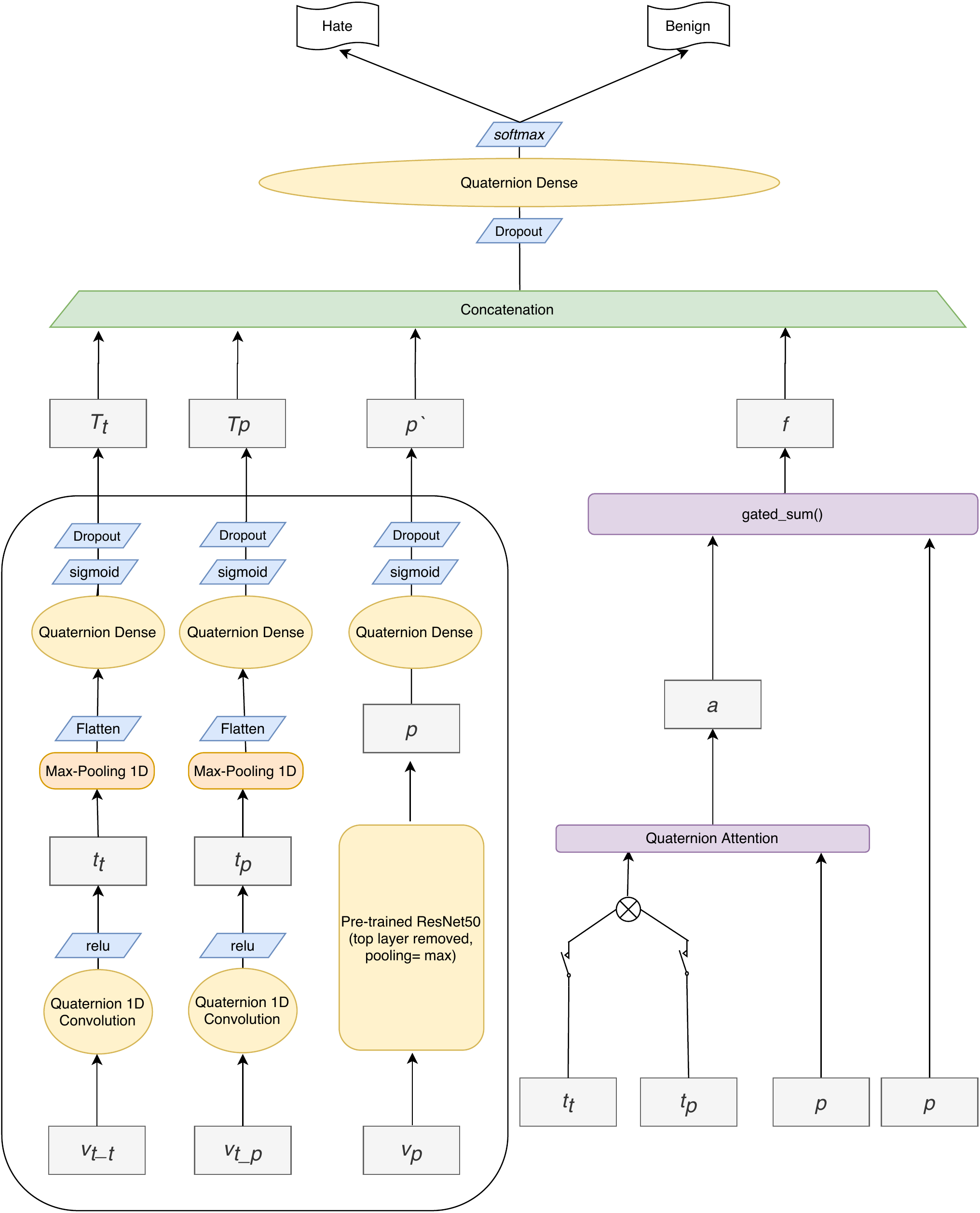}}
\caption{
Quaternion fusion model\\
} 
\label{fig1}
\end{figure}

\section{Methodology}
\label{sec:model}
We propose QUARC-a QUaternion multi-modal fusion ARChitecture, which uses the QCNN \cite{ref_article4} as the baseline architecture for all text, image and fusion part of model. Overall structure of our model is shown in \hyperref[fig1]{Figure $1$}.

\subsection{Text}
\label{ssec:3.1}
This text model is inspired by \cite{ref_article11}, where CNN was used for sentence classification. There are two types of text, tweet text and image text. Text model is same for both but are trained separately as follows. 
\begin{enumerate}
    \item \label{sssec:3.1.1} Firstly, to process the text before generating the word embeddings, we proceed as follows:
    \begin{itemize}
        \item We removed URLs and associated symbols.
        \item as it contain considerable amount of emojis, they are changed in text.
    \end{itemize} 
    \item \label{sssec:3.1.2}Then we retrieve pre-trained embeddings from GloVe\cite{ref_article12} for tweet as well as image texts. Words not present in GloVe embeddings are set randomly. As data-set is not large enough, we don't train  GloVe  word  embedding  model\cite{ref_article2}. We get vectors $v_{t\_t}$ \&  $v_{t\_p}$ for tweet text and image text respectively after proper padding with word limit to be 150 per input text.   
    \item \label{sssec:3.1.3}Then we apply quaternion $1$-D convolution layer with same padding to keep input and output length same for applying quaternion attention in additional fusion.
    \item \label{sssec:3.1.4}After that we apply max-pooling, ReLu and flatten it.
    \item \label{sssec:3.1.5}Lastly, the resulting vector representation $T_{t}$ (tweet text) \& $T_{p}$ (image text) (\hyperref[fig1]{see Fig.1}) obtained after applying quaternion multi-layer perceptron (Q-MLP) and a dropout layer, is used in final concatenation.
\end{enumerate}

\begin{table}[t]
\centering
\begin{tabular}{ | l | l | } 
 \hline
 \textbf{Hyper-parameters} & \textbf{Values} \\
 \hline
 Dropout rate & $0.35$  \\ 
 \hline
 Batch Normalization & Quaternion Batch Normalization \\
 \hline
 Kernel Initializer & Glorot \\
 \hline
 Optimizer & Adam \\
 \hline
 Loss Function & Binary Cross-entropy \\
 \hline
 Epochs & $20$ \\
 \hline
 Batch Size & $128$ \\
 \hline
\end{tabular}
\caption{Hyper-parameters}
\label{tab:hyperparameter_label}
\end{table}

\subsection{Image}
\label{ssec:3.2}
The image model is trained as follows:

\begin{enumerate}
    \item \label{sssec:3.2.1} First we resize the images in $32 \times 32$ pixels. Now we make sure all images are 3-channeled, if not, we convert them into it by adding null channels.
    \item \label{sssec:3.2.2} These input images are passed through pre-trained ResNet50 without its output layer.
\begin{figure}[htbp]
\centerline{     
\subfigure[``What separates humans from animals?"]{\label{fig:a}\includegraphics[width=30mm]{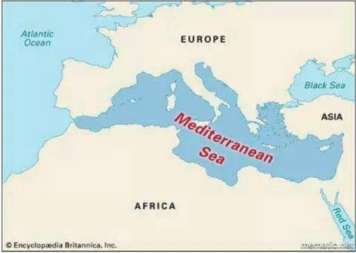}}
\subfigure[``It's so hard being a nigga"]{\label{fig:c}\includegraphics[width=20mm]{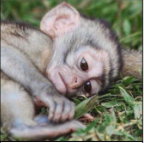}}
\subfigure[``Hey.... fucck raghead"]{\label{fig:d}\includegraphics[width=25mm]{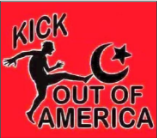}}
}
\caption{Sample tweets with caption as their tweet texts. \cite{ref_article2}}
\label{fig 9}
\end{figure}

    \item \label{sssec:3.2.3} The vector ($p`$) finally obtained after passing the max pooled output ($p$) through quaternion dense layer and dropout layer is used in final concatenation.
\end{enumerate}

\subsection{Fusion}
\label{ssec:3.3}
\subsubsection{Simple fusion}
\label{sssec:3.3.1}
Here we simply concatenate $T_{t}$, $T_{p}$ and $p`$. Then we apply  dropout, MLP and softmax on $concat$($T_{t}$, $T_{p}$, $p`$) for the final hate speech classification.

\subsubsection{Symmetric gated fusion}
\label{sssec:3.3.2}
We follow the symmetric gated summation approach similar to  \cite{ref_article13} and \cite{ref_article5}. In $gated\_sum(a,p)$, we linearly transform the input vectors $a$ \& $p$ such that they both have the same dimension. Then, we calculate $\beta_{a}, \beta_{p}$ \& $m$ as given in the equ. $(\ref{eq 1})$, $(\ref{eq 2})$ \& $(\ref{eq 3})$, where the weight \& bias vectors ($W_{a}, U_{a}, W_{p}, U_{p}, W_{m}, U_{m}, B_{a}, B_{p}, B_{m}$) are to be learned by the model. The $visual\; modulation \; gate$ ($f$), as referred by \cite{ref_article13}, can dynamically control the combination of text and image information.         

\[a'=W_{a}\cdot a+b_{a}\]
\[p'=W_{p}\cdot p+b_{p}\]
\begin{equation} \label{eq 1}
    \beta_{a}= \sigma(W_{a}\cdot a'+U_{a}\cdot p'+B_{a})
\end{equation}
\begin{equation} \label{eq 2}
    \beta_{p}= \sigma(W_{p}\cdot a'+U_{p}\cdot p'+B_{p})
\end{equation}
\begin{equation} \label{eq 3}
    m= \tanh{(W_{m}\ast a'+U_{m}\ast  p'+B_{m})}
\end{equation}
\begin{equation} \label{eq 4}
    gated\_sum(a,p)= f= \beta_{a}\ast a'+\beta_{p}\ast m
\end{equation}

Using the quaternion attention architecture (\cite{ref_article3}), we try to find how much the given photo information (query vector) is related to the certain part of the text (context vectors). The explicit equation for the attention is given in the equ. $(\ref{eq 5})$. The weighted sum $a_{c}$ of the text vectors, using the photo information, is then passed through the above $gated.sum()$ function along with the image vector $p$. 
\begin{equation} \label{eq 5}
s_{c_i}= softmax(c^T_{i}\cdot W_{a}\otimes p') \;\;\;\; i= 1,...,n 
\end{equation}
\begin{equation} \label{eq 6}
a_c= \sum (s_{c_i}\ast c_{i})
\end{equation}
where\;$c= [c_{1},.., c_{n}]\;$ is the text ($t_p$ = image text / $t_t$= tweet text) vector obtained in
\hyperref[sssec:3.1.3]{3.1.3}. 

The resultant vectors $a_{t_t}$, $a_{t_p}$, $a_{t_w}$ are then passed through the $gated\_sum()$ along with the image vector $p$ (where, $a_{t_w}= a_{t_{t}} +  a_{t_{p}}$).
We now concatenate the vectors $T_{t}$, $T_{p}$, $p`$, $gated\_sum(a_{t_t},p)$, $gated\_sum(a_{t_p},p)$\& $gated\_sum(a_{t_w},p)$ and apply QMLP, dropout \& softmax for hate speech classification.

\subsection{Models}
\label{ssec:3.4}
So in summary, here are the fusion models with left side of equality denoting model with its input and right side of equality giving concatenation layer with its input which is then followed by dropout, quaternion dense layer and softmax operations for the final hate speech classification.
\begin{IEEEeqnarray*}{lCr}
 \textbf{Model\; 1} : \\
 F(v_{t_t}, v_{t_p}, v_p) = C(T_t,  T_p, p`, GS(a_{t_t}, p), GS(a_{t_p}, p), GS(a_{t_w}, p)) \\
 \textbf{Model\; 2}: F(v_{t_t}, v_p) = C(T_t, T_p, p`, GS(a_{t_t},p)) \\
 \textbf{Model\; 3} : F(v_{t_p}, v_p) = C(T_t, T_p, p`, GS(a_{t_p},p) \\ 
 \textbf{Model\; 4} : F(sum(v_{t_t}, v_{t_p}), v_p) = C(T_t, T_p, p`, GS(a_{t_w},p)) \\
 \textbf{Model\; 5} : simpleconcat(v_{t_t}, v_{t_p}, v_p) = C(T_t, T_p, p`) \\
 \textbf{Model\; 6} :v_{t_t} = C(T_t) \\
 \textbf{Model\; 7} :v_p = C(p`)
\end{IEEEeqnarray*}
Here \textit{F} represents \textit{fusion}, \textit{C} represents \textit{concat}, \textit{GS} represents \textit{gated\_sum}

\begin{table*}[t]
\centering
\captionsetup{justification=centering}
\begin{tabular}{ |r|l|r|r|r|r|r|r|r| } 
\hline
\textbf{Model No.} & \textbf{Model} &  \multicolumn{2}{c|}{\textbf{\# Trainable Param.}}  & \multicolumn{2}{c|}{\textbf{ROC-AUC}} &  \multicolumn{2}{c|}{\textbf{Avg Precision Score}} \\ 
\hline
& &  \multicolumn{1}{c|}{$\rm I\!R$} & \multicolumn{1}{c|}{$\rm I\!H$} & \multicolumn{1}{c|}{$\rm I\!R$} & \multicolumn{1}{c|}{$\rm I\!H$} & \multicolumn{1}{c|}{$\rm I\!R$} & \multicolumn{1}{c|}{$\rm I\!H$}  \\ 
\hline
1. & $fusion(v_{t_t}, v_{t_p}, p)$ & $2,810,573$ & $719,837$ &70.28  &71.49  &   $0.8664$ & $0.8763$ \\ 
\hline
2. & $fusion(v_{t_t}, p)$ &$1,241,545$  & $311,353$ &70.39 &71.47 &  $0.8754$ & $0.8764$ \\ 
\hline
3. & $fusion(v_{t_p}, p)$ & $1,241,545$  & $311,353$ &70.46 &70.80 &  $0.8717$ & $0.8746$ \\
\hline
4. & $fusion(sum(v_{t_t}, v_{t_p}), p)$ & $2,752,305$ & $701,553$ &70.03 &70.79 &  $0.8679$ & $0.8775$ \\
\hline
5. & $simpleconcat(v_{t_t}, v_{t_p}, p)$  & $1,219,894$ & $301,361$ &70.84 &70.87 &  $0.8671$ & $0.8748$ \\
\hline
6. & $v_{t_t}$  & $530,229$ & $132,729$ &70.80 &71.38 &  $0.8729$ & $0.8738$ \\
\hline
7. & $v_p$  & $131,201$ & $32,897$ &50.34 &51.28 &  $0.7436$ & $0.7474$ \\
\hline
\end{tabular}
\caption{Performance of the proposed models}
\label{tab:result_label}
\end{table*}

\section{Experiments}
\label{sec:4}

\subsection{Data}
\label{ssec:4.1}
We did our experiment on the only publicly available dataset having both image and text that we could gather, which is MMHS150K \cite{ref_article2} twitter data-set for hate speech. It is a manually annotated data comprising $150,000$ twitter tweets, each containing a tweet text component (may contain emojis, and URL), a tweet image component and may also contain image text component (text extracted from tweet image using Google Vision API). \hyperref[fig 9]{Figure $2$}, shows few sample images with corresponding tweet text as its caption. Out of this validation set contains $5,000$ and test set $10,000$ tweets. 
Our code is available at  \url{ https://github.com/smlab-niser/quaternionFusion}.

\subsection{Hyper-parameters}
\label{ssec:4.3}
We use 100-dimensional twitter GloVe embeddings, which are trained on $2$B tweets, making it suitable for our experiment. Maximum word length is set to $150$ for each input. We use $1$D-convolution windows of size $5$ with $128$ filters for both image and tweet text while $2$D-convolution windows of 

\begin{figure}[htbp]
\centerline{
\subfigure[Trainable Parameters]{\label{fig:a}\includegraphics[width=0.5\linewidth]{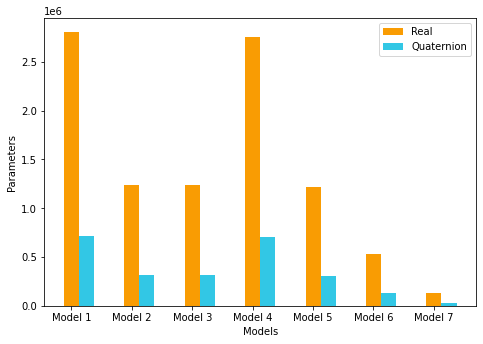}}
\subfigure[Avg. Precision Score]{\label{fig:c}\includegraphics[width=0.5\linewidth]{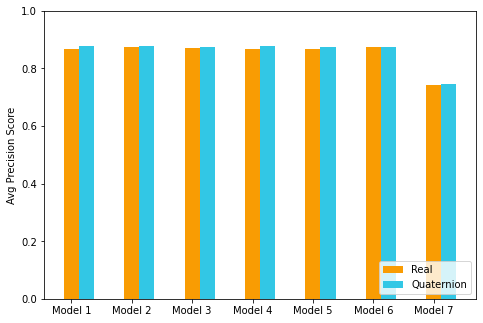}}
}
\centerline{
}
\caption{Performance of models. For model no. \hyperref[ssec:3.4]{see Sec. \RN{3}.D}}
\label{fig 3}
\end{figure}

size $2\times2$ \& $3\times3$ for image training. Dimensions of $T_{t}$, $T_{p}$ \& $p$ is set to $64$. Other hyper-parameters are given in \hyperref[tab:hyperparameter_label]{Table $1$}.

\subsection{Results \& Analysis}
\label{ssec:4.4}
Results of our experiments are given in \hyperref[tab:result_label]{Table \RN{2}}. In this table, we have real and quaternion version of each model. The table shows number of trainable parameters, ROC-AUC score and average precision score for each fusion model.

In the \hyperref[tab:result_label]{Table \RN{2}}, we can see that quaternion version for each fusion model shows almost $75\%$ reduction in trainable parameters as compared to the real one. Despite this reduction in parameters, we can see comparable performance (ROC-AUC score and average precision score) between quaternion and real. The quaternion version of $fusion(v_{t_t}, v_{t_p}, p)$ model shows best ROC-AUC score and the quaternion version of $fusion(sum(v_{t_t}, v_{t_p}), p)$ model shows best average precision score among all models.

The ROC-AUC score of $v_{t_t}$ model is slightly better than some of the fusion models, which is against our expectation. However, similar trend is seen in \cite{ref_article2}. Also, we traded off performance by resizing the images in $32 \times 32$ pixels (while \cite{ref_article2} 
considered images of 500 pixels) because otherwise matrices corresponding to images were getting too big to be handled by our current setup. We suspect this might be the reason for information loss in image, hence, contributing less in the fusion models.

From the results of \cite{ref_article2}, we can observe that performance isn't improving much from only text ($T_{t}$) to fusion of text and image text ($T_{p}$) and finally fusion of all three, regarding which \cite{ref_article2} suspects that fusion model is not able to extract information other that the text modality. However, we suspect that this can be asserted to the unique nature of this dataset and its noise because of extracted image text and the subjectivity of annotator (also mentioned by \cite{ref_article2}). But, the improvement of performance with fusion can be seen in experiments of  \cite{ref_article5}, which is done over Facebook dataset (not public). 

\section{Conclusion}
\label{sec:5}
Social media is an important platform to bring our society together with healthy discourse. But, because of some bad apples, even this place is becoming toxic, unsafe and, a propaganda tool. Hate speech nowadays comes in the combination of image, video, audio and text. So multi-modal models become essential for hate speech classification. However, these models are having high number of parameters. In this paper we conclude that quaternion based multi-modal fusion model not only reduces parameters in comparison to its real counterpart but also achieves comparable performance because of the ability of quaternion to handle images better. These results motivate us to further look into similar model based on another such hyper-complex space, octonion algebra, an extension of quaternion algebra. Also, improvements can be made by considering video as part of posts. Further we are working on developing Qu-BERT architecture for multi-modal fusion.

\end{document}